\pdfoutput=1
\documentclass[a4paper,12pt]{article}
\usepackage{graphicx} 
\usepackage{url} 
\usepackage{authblk}

\PassOptionsToPackage{numbers, compress}{natbib}

\usepackage[utf8]{inputenc} 
\usepackage[T1]{fontenc}    
\usepackage{hyperref}       
\usepackage{url}            
\usepackage{booktabs}       
\usepackage{nicefrac}       
\usepackage{microtype}      
\usepackage{xcolor}         

\usepackage{latexsym}
\usepackage{inconsolata}
\usepackage{multirow}
\usepackage{bbding}

\usepackage{comment}
\usepackage{amsmath,amssymb,amsfonts}

\usepackage[ruled]{algorithm}
\usepackage{algpseudocode}
\usepackage{caption}
\usepackage{xfrac}

\usepackage{comment}
\usepackage[capitalize]{cleveref}
\crefname{section}{Sec.}{Secs.}
\Crefname{section}{Section}{Sections}
\Crefname{table}{Table}{Tables}
\crefname{table}{Tab.}{Tabs.}
\let\oldnl\nl
\newcommand{\nonl}{\renewcommand{\nl}{\let\nl\oldnl}}

\title{Flexible and Effective Mixing of  Large Language Models into a Mixture of Domain Experts}

\author[1]{Rhui Dih Lee}
\author[2]{Laura Wynter}
\author[3]{Raghu Kiran Ganti}
\affil[1]{IBM Research, Singapore, rhui.dih.lee@ibm.com}
\affil[2]{IBM Research, Singapore, lwynter@sg.ibm.com}
\affil[3]{IBM Research, Yorktown Heights, NY, USA, rganti@us.ibm.com}

\begin{document}
\maketitle

\begin{abstract}
We present a toolkit for creating  low-cost Mixture-of-Domain-Experts (MOE) from trained models. The toolkit can be used for creating a mixture from  models or from adapters. We perform extensive tests and offer guidance on defining the architecture of the resulting MOE using the toolkit. A public repository is available at \url{https://github.com/RhuiDih/moetify}.

\end{abstract}

\section{Introduction}

Mixture of Experts (MOE) models, like Mixtral, have been shown to perform very well, often better, than larger, dense models like LLaMa-70b \cite{mixtral,HFMOE,deepseek}. In addition, MOE models activate fewer parameters for each token than dense models, and hence can offer faster inference response times. 
These two attributes have contributed in making MOE models very popular. In fact, while its architecture is not known publicly, GPT-4 \cite{gpt4} is rumoured to be a sparsely-activated (e.g. 2-activated experts per-token)  MOE model, where that design choice, if accurate, may have been made to get a larger capacity with smaller inference cost due to the high number of inference calls that GPT-4 sees.

Given the nature of an MOE model, composed of several “experts,” it is tempting to cast the expert sub-models as each having an application domain expertise. In the well-known MOE models, though, this is not the case: the “experts” are not specialised during training.  The architecture of an MOE is designed so that a gating function, also known as a  router, decides which experts  handle each token. However, the routers and the expert modules  are trained simultaneously.
While it is in theory possible to modify the training of an MOE model so that the MOE expert modules comprise real “experts”, the training process would be lengthy and difficult, and may not deliver the desired outcome. 
The creation of MOE models can, however, be achieved at nearly-zero cost by mixing multiple trained, and fine-tuned, models. 
Furthermore, by mixing multiple trained models into an MOE, one can select the trained models so that they represent domain experts.

We wish to enable rapid and low-cost MOE model creation to augment the capabilities of a given source model of interest. One needs only to select additional models with the same architecture as the source model as experts, and then combine the trained expert models with the source model of interest into an MOE. By selecting domain-specialised, trained models of interest   to augment the capabilities of the source model, the resulting MOE model can deliver the promise of a true Mixture of Domain Experts. 

We provide a general-purpose toolkit for using trained models in a Mixture of Domain Experts MOE with a focus on the flexibility offered. While a router, or gate, can be trained on a small amount of relevant data to improve the performance of the Mixture of Domain Experts MOE, we find that it is not always necessary. Hence our toolkit offers the flexibility to create a Mixture of Domain Experts MOE in multiple ways including without router training. When there are only a few high-quality  experts, our  Gate-less MOE architecture can be the best solution. We find that the Gate-less architecture is  competitive with and can even outperform  router-based architectures yet is cheaper to produce.   To  reduce the inference cost of the Gate-less MOE when the number of expert modules increases, we also propose a Noisy MOE that performs nearly as well as the Gate-less MOE, does not require training, and offers a lower inference cost. In addition, our toolkit  offers the capability to train the router or train a combination of the router and the embedding layers. We also offer the possibility to create the MOE from trained LoRA adapters.

We cover in Section \ref{sec:relatedwork} the related work including other libraries offering a similar functionality. In Section \ref{sec:method}, we describe the methodology proposed and implemented in our open source repository. Section \ref{sec:results} provides detailed experimental results using the proposed methodology. In particular, we justify the development of the Gate-less and the Noisy MOE and contrast it with the approaches provided in other related works. Section \ref{sec:conc} concludes the paper. The open-source  library  can be found at \url{https://github.com/RhuiDih/moetify}.

 \section{Related work}
 \label{sec:relatedwork}
 
Evidence, such as in \cite{cheng2024adapting}, shows that models specialised, through fine-tuning, to a particular domain outperform generalist models on their domains of interest. In cases where an MOE model comprises multiple domain-specialised expert models, it was shown in \cite{FAIR} that a mixture-of-multiple-experts model can outperform their respective source expert models.  
 
Taking it one step further, we can imagine a set of trained models, each having a skill in a particular domain. Each MOE can mix a targeted subset of skill-based models to satisfy the distinct needs of each individual user. Given the exceptionally low cost of creating these mixed MOE models, they can be customised rapidly on demand, for each use, with only the skills of interest. 

There have been a few other efforts to enable mixture of experts model creation from trained models, the first of which is due to Charles Goddard \cite{goddard, mergekit} who created the Mergekit repository. The recommendation provided there is to set the router weights from the  hidden states in the FFN of each expert obtained when running each expert on a set of targeted prompts. Specifically, the author suggests compiling of a list of positive and negative prompts and then using a provided script which combines their hidden state representations by averaging and normalizing them, for each expert. 

The Mergekit library was used to create a series of MOE models documented in a Hugging Face blog article \cite{maxime} which includes  numerical results with the resulting MOE models. While  the MOE performs  well, it does not, on the vast majority of the tests used for evaluation, perform better than its  expert models. In particular, the MOE performs worse than the model used as the  base and as one of the experts on most of the tasks used to evaluate it. However, the same author posted an earlier mixed MOE that did outperform its constituent models \cite{maximephi} though it was not included in the blog article \cite{maxime}. Later, a similar library \cite{mergoo} was created by a different team, however, no experimental results were provided to demonstrate if or how well the resulting model works. 

The authors of \cite{selfmoe} propose a similar approach but require the experts to be LoRA adapters and the use of a single linear-layer router shared across all of the LoRA layers. Code is not provided.  We note that a LoRA-adapter based architecture can be achieved with our methods and toolkit,  along with further flexibility that we provide in the definition of the experts and the routing mechanism. We evaluate an MOE variant similar to that of \cite{selfmoe} in our experiments.

In \cite{austria} the authors propose an "on-demand selection and combination" of LoRA adapters at inference time and provide a their code publicly. Their method consists of a scoring strategy to identify the top $K$ adapters and various weighting (and parameter averaging and ensembling) strategies for combining the adapters. However, their results are limited to comparing perplexity  across the weighting strategies. In addition, their scoring methods are not practical, as they require  in general each adapter's  "training domain dataset" to evaluate proximity to the input.

Lastly, we mention CALM \cite{calm}, a   version of the MOE idea comprising  two models: a base model and an augmenting model. The approach requires learning, for selected layers,  linear transformations
from the augmenting model to the base model and  a cross-attention layer between each projected layer and the corresponding base layer. However, in their paper, the composition of the base model with a small augmenting model is almost always inferior to fine-tuning the base model on the same dataset used for the composition training. While the authors claim that the method can extend to more than one augmenting model, inference would be far too costly when multiple experts are used as the forward pass makes use of the full models of the base and the experts.

 \section{Augmenting  an LLM with other expert LLMs}
\label{sec:method}
 
We are interested in augmenting the capabilities of a large language model to improve its performance on multiple, related domains, and to do so at a low computational cost. When one has available pre-trained, fine-tuned domain expert models, as is the case on the Hugging Face Model Hub\cite{HFhub}, augmenting a given model to address multiple, related domains becomes an appealing and feasible task. 

\begin{figure}[ht]
\centering
\includegraphics[width=\linewidth]{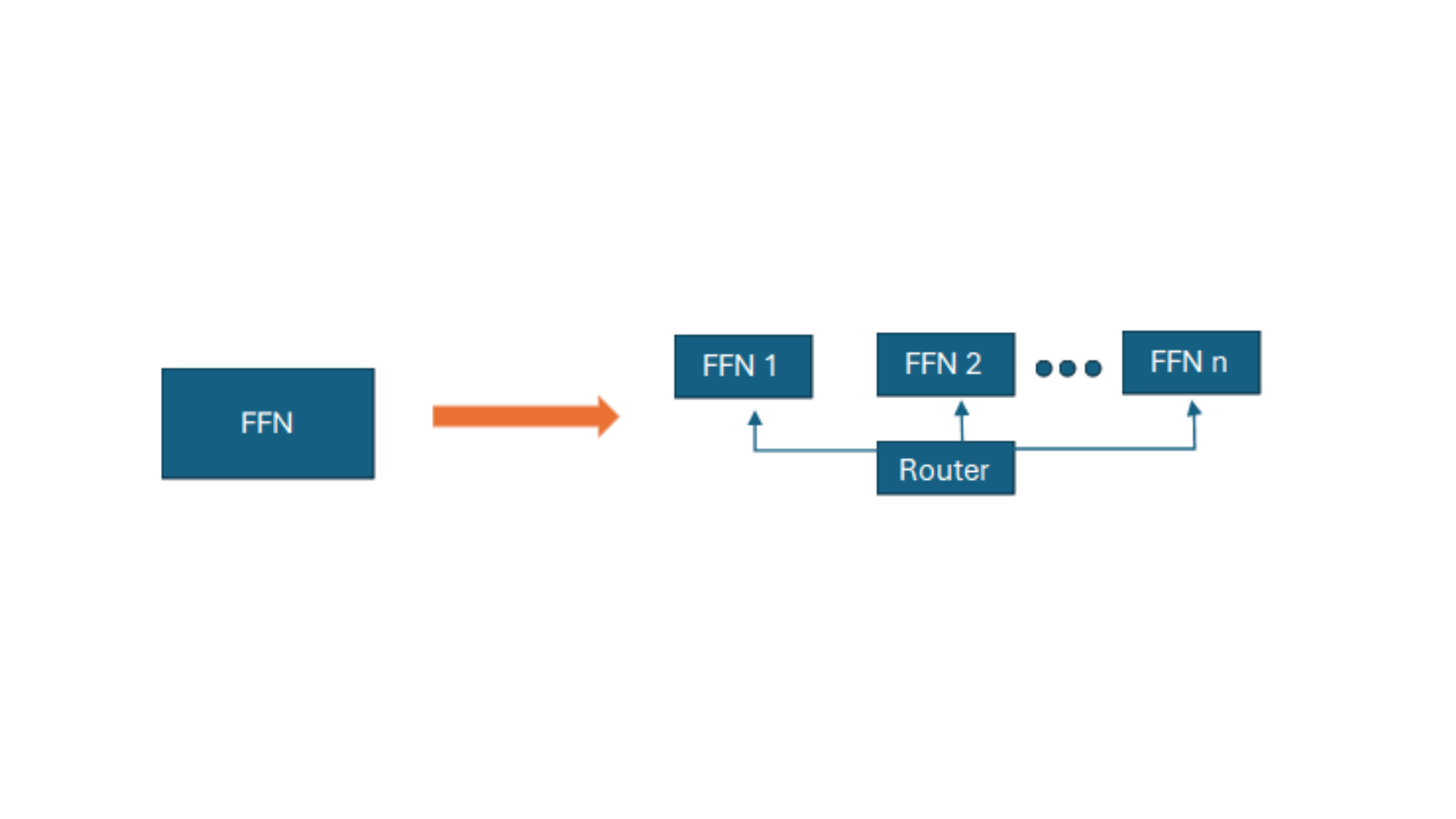}
\caption{Illustration of MOE model  swapping a its FFN with a set of FFN layers and a router }
\label{fig:model-mixing}
\end{figure}

Our MOE Model Mixing toolkit swaps the FFN layers of each expert model, along with a gate, in place of the FFN layers of a base model.  See Figure 1. 
In \cite{mergekit}, it was suggested to set the router parameters as the hidden state representations for each expert. Specifically, it was suggested  that  passing  a  set of positive and negative prompts through each expert model, and then averaging and normalising the hidden states thus obtained, the resulting values can be used as the router weights of each expert.  
 We found that using this type of hidden representation in the gate does not work well.  
The authors of \cite{mergoo} assume that users of a mixed MOE will fully fine-tune the resulting MOE. This, we   found, is not needed. We  suspect that  fine-tuning the mixed MOE for a few epochs results in the MOE losing the ability of its expert modules to handle well the domains that each expert was trained on.  

We aim to leverage well-trained and effective expert modules and use them all as first-class citizens. We thus propose  creating a Gate-less MOE, which assigns an equal weight to each expert. We show that when the number of expert models is small, this can be an optimal strategy for MOE model mixing in terms both of model creation cost and subsequent model performance on evaluation tasks. 
 However, when the number of expert models grows, one may wish to avail of the sparsity that a top-$K$ strategy affords, whereby only the top $K$ expert modules are activated for each token. 
 To this end, we also allow creating a Noisy MOE, which uses a single linear layer of dimension hidden-size by number of experts, and defines the weight of each element to be white noise centered at 0 with a small variance, and then uses a top $K$ strategy to select the $K$ experts to route each token to. We show in Section \ref{sec:results} that this Noisy MOE  works almost as well as the Gate-less MOE  and provides faster inference time when there are more than 2 experts. 
 
 The flexibility of the methods we provide means that experts can be readily swapped in and out,  with both the Gate-less MOE and the Noisy MOE, at practically zero cost and with no training required.
 For the Noisy MOE with top-$K$ routing, an  option of specifying an “always-on” expert is provided. Optional attention-layer mixing, along with the FFN mixing, is also supported. 
 While we have found that the routers need not be trained to achieve good results, our toolkit offers the possibility to train the routers or a combination of the routers and the attention layers. In addition, while our results show that using the FFN layers of the experts is generally preferred, we also enable creating an MOE from LoRA adapter experts.

\section{Experimental results}
\label{sec:results}

We wish to extend the capabilities of   a model using additional pre-trained and tuned expert models.  First, we demonstrate this using the InstructLab-enhanced Merlinite model released by IBM instructlab/merlinite-7b-lab \cite{instructlab}   as our base model as well as one of our experts. 
We  make use the following additional models as  experts, sourced from Hugging Face Model Hub \cite{HFhub}:  openchat /openchat-3.5-1210 \cite{openchat}, NousResearch/Hermes-2-Pro-Mistral-7B \cite{hermes}, and meta-math/ MetaMath-Mistral-7B \cite{metamath}. All three additional expert models have permissible usage licenses and, based on their performance, should enable merlinite to perform better in math and general language capabilities.
Later, in Section \ref{sec:llama3}, we perform ablation studies on llama3-8B models and also  compare against a low-cost adapter-based MOE, including an approach similar to that of   \cite{selfmoe}.

Our evaluations  are performed using TruthfulQA \cite{truthfulqa}; AGIEval \cite{agieval}; GPT4ALL \cite{gpt4all}, comprising 
piqa, openbookqa, boolq, arceasy, arcchallenge, winogrande, and hellaswag; MMLU 5-shot \cite{mmlu}; BBH-few shot \cite{bbh}; the math domain, comprising MathQA \cite{mathqa} and GSM8K 5-shot \cite{gsm8k}; and the Finance domain \cite{fin}, comprising convfinqa,	finheadline,	fiqa, and	fpb. 

When the routers are trained, the hyperparameters used were set to num\_train\_epochs = 1, per\_device\_train\_batch\_size = 1, gradient~\_accumulation~\_steps = 16, learning\_rate=1e-04, and lr\_scheduler\_type as constant. For router training,   instruction tuning is performed using a  dataset combining 50K  examples each from metamath \cite{metamath}, pubmedQA \cite{pubmedqa} and FLAN \cite{flan}. When performing extended pre-training of the routers, we  combine 50K  examples each from wikidump \cite{wikidump} and open-web-math \cite{openwebmath}. The 2x MOE in the next section mixes Merlinite with MetaMath.

\begin{figure}[ht]
\centering
\includegraphics[width=\linewidth]{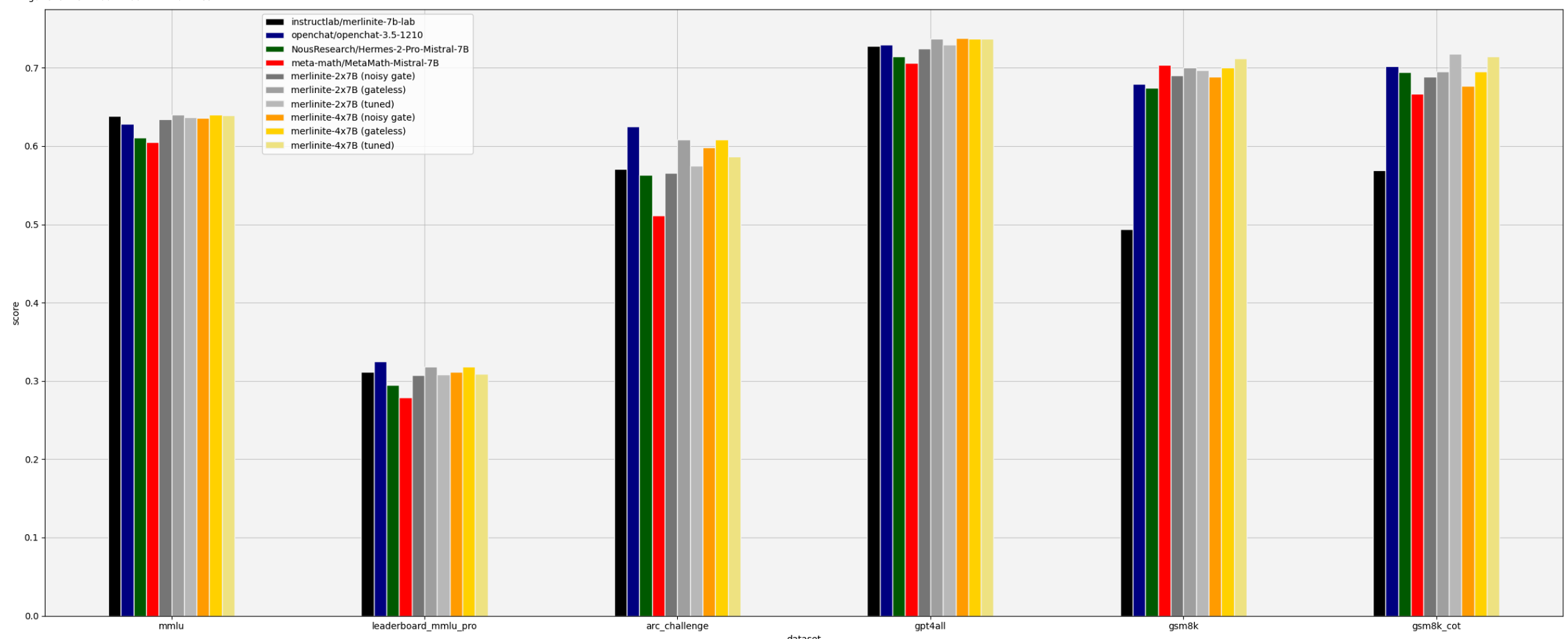}
\caption{Main results  of MOE creation on Merlinite with 2 and 4 experts. The first four bars in each section provide evaluations with the trained expert models individually. The set of three grey bars next from the left are the 2x MOE, and the three shades of yellow bars are the 4x MOE. The overall observation is that the MOEs are all very competitive, with or without router training and with both 2 and 4 experts. Router training is primarily  advantageous on the math tasks, GSK8K-COT in particular. }
\label{fig:results}
\end{figure}

\subsection{Low-cost MOE creation is a viable approach}

The main results using the Merlinite model are provided in  Figure \ref{fig:results}. The key observation is that low-cost creation of an MOE from trained expert models is a viable approach to improving the performance of a model in a cost-effective manner. The first four bars from the left in each section of the figure show the evaluation results with the expert models individually. The different sections are different evaluation tests with general knowledge and reasoning tests MMLU, MMLU-pro, ARC-challenge and GPT4All followed by two math test sets, GSM8K and GSM8K-COT. While the original Merlinite model performs well on MMLU and GPT4All, it is clear that its performance is lacking on the other evaluation tests. We aim therefore to complement Merlinite's performance on other tasks without degrading its performance on MMLU and GPT4All. Figure \ref{fig:results} shows that this is indeed achievable. Focusing here on the first 2 grey bars and the first 2 yellow bars from the left, in each section, we see that Merlinite's performance is maintained where it was good and is improved considerably where it was lacking. While there are differences between the 4x MOE and the 2x MOE, both are competitive. 

The Noisy MOE policy, which chooses the top $K=2$ experts for each token  with a randomised, white-noise-based policy, shows a slight degradation as compared to the Gate-less MOE but still performs very well. 
With 4 expert modules, the inference cost of  Gate-free MOE remains modestly higher than the original source model, merlinite. See Table \ref{tab:inftime}. However, with a larger number of experts, activating all  would be undesirable from an inference cost perspective.  Noisy MOE provides performance nearly as good as  Gate-free MOE with a lower inference cost.   Inference times for the Gate-less MOE and Noisy MOE on a single A100-80GB GPU are provided in Table \ref{tab:inftime} and memory usage in Table \ref{tab:mem}. Memory usage  is identical for both versions of the MOE, leading to an out-of-memory error when 8 experts are used. Note that in the 2x MOE setting,  Gate-free MOE is preferable, as both experts are activated using both policies and Gate-free MOE does not require a linear router layer, thereby saving  inference cost.

\begin{table}[!ht]
    \centering
    \begin{tabular}{|l|l|l|l|l|}
    \hline
        Inference time (s) & 2X   & 4X  & 6X & 8X  \\ \hline\hline
        Gate-free & 1.079 & 
        1.888 & 2.694 & OOM \\ \hline
        Noisy MOE & 1.155 & 
        1.179 &
        1.203 & OOM \\ \hline
    \end{tabular}
    \caption{Inference time on an A100-80GB GPU for Merlinite Gate-free MOE and Noisy MOE for 2,4, 6, and 8 experts (denoted 8X, 6X, 4X,  and 2X). Noisy Gate MOE uses a Top-$K$ policy with $K=2$. OOM means 'out-of-memory'. }
    \label{tab:inftime}
\end{table}

\begin{table}[!ht]
    \centering
    \begin{tabular}{|l|l|l|l|l|}
    \hline
        Memory (GB) & 2X   & 4X  & 6X & 8X  \\ \hline\hline
        Gate-free & 27.137 & 
        48.642 & 70.146 & OOM \\ \hline
        Noisy MOE & 27.137 & 
        48.641 &
       70.145 & OOM \\ \hline
    \end{tabular}
    \caption{Memory utilisation on an A100-80GB GPU for Merlinite Gate-free MOE and Noisy MOE for 2,4, 6, and 8 experts (denoted 8X, 6X, 4X,  and 2X). Noisy Gate MOE uses a Top-$K$ policy with $K=2$. OOM means 'out-of-memory'. }
    \label{tab:mem}
\end{table}

\subsection{Router training can be beneficial but is not required }

 We consider several  paradigms for training the router, including extended pre-training, instruct-tuning of the router and instruct-tuning of both the router and o-projection layers.  We find that the decrease in loss is moderate during router training, implying that  the ability of the router to learn is  somewhat limited. 
 We conjecture that the capacity of the gate can be insufficient to learn a complex routing policy with a small or moderate amount of data. Furthermore,  we observe that, in many cases, router training is simply not necessary to achieve good performance from the mixed MOE.

\begin{figure}[ht]
\centering
\includegraphics[width=\linewidth]{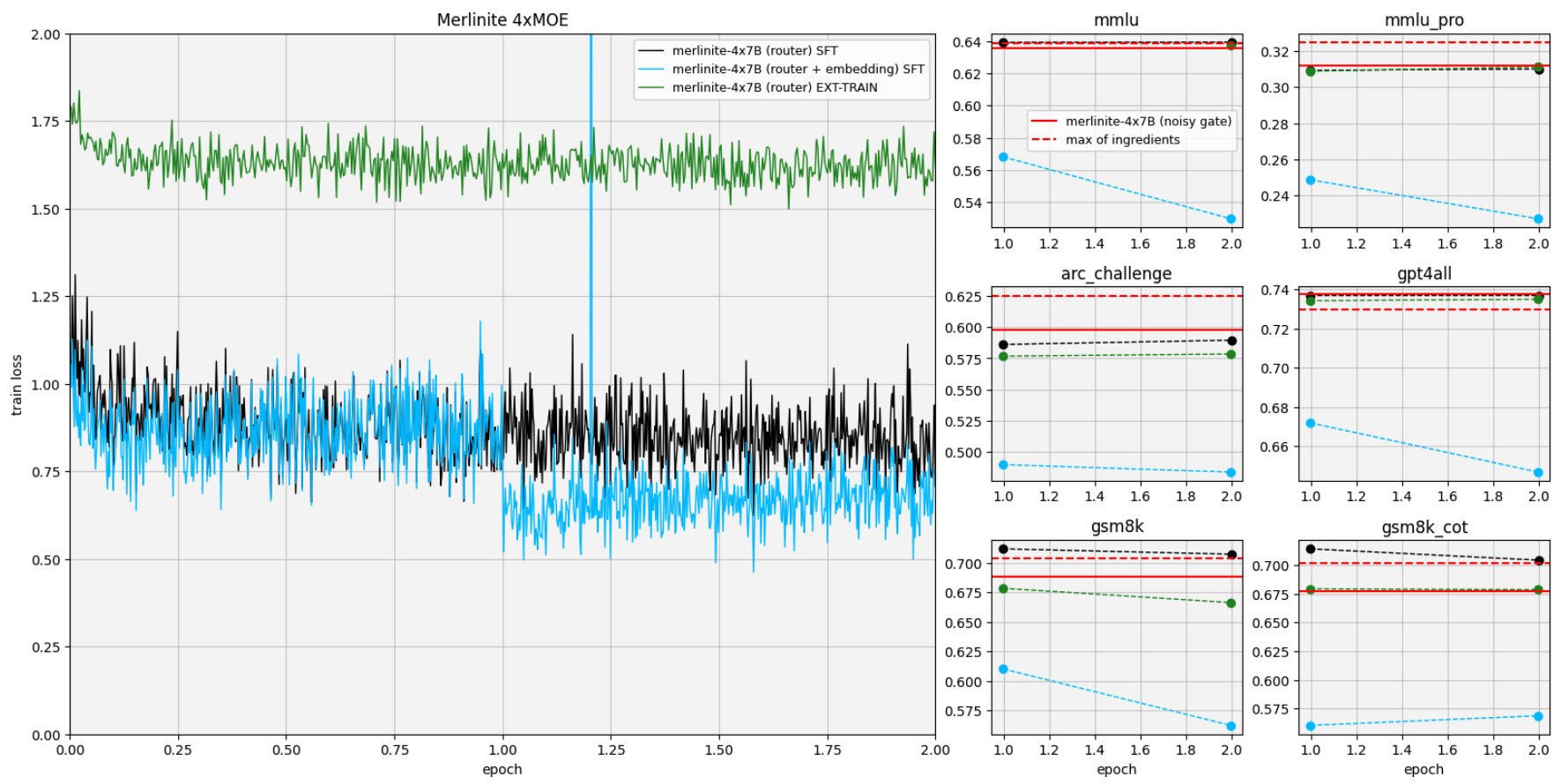}
\caption{Illustration of training loss of the Merlinite 4X  MOE models with 3 variants of the training paradigm: instruction tuning of the router and router with embedding layer, and extended pre-training of the router. On the right side we show evaluation results using the checkpoints after 1 and 2 epochs of training.  The Noisy MOE (no training) evaluation result is the solid red line while the best result of the 4 experts alone  is the dashed red line. We see that router training offers benefit    on the math tasks. }
\label{fig:4xloss}
\end{figure}

\begin{figure}[ht]
\centering
\includegraphics[width=\linewidth]{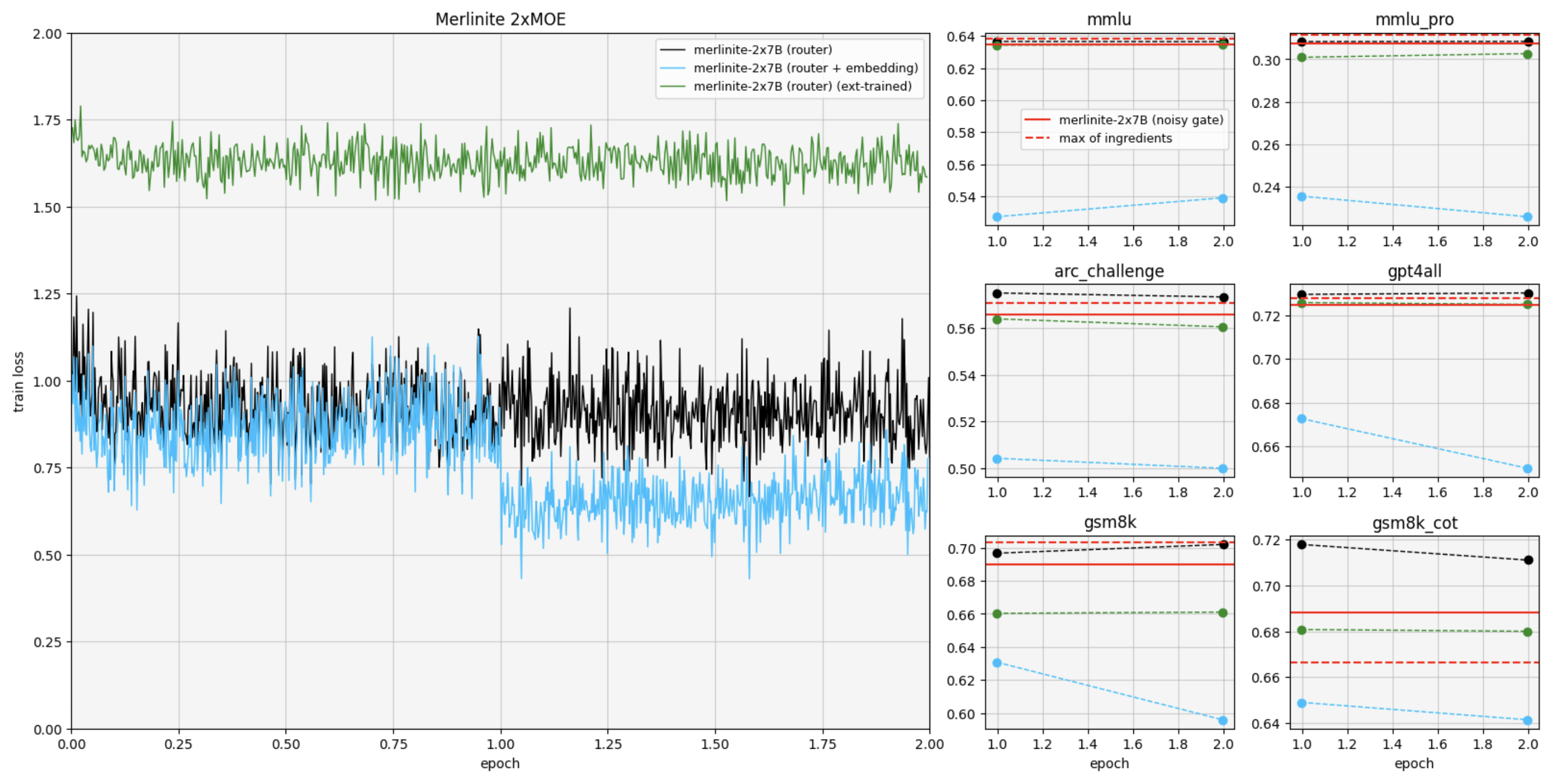}
\caption{Illustration of training loss of the Merlinite 2X   MOE models with 3 variants of the training paradigm: instruction tuning of the router and router with embedding layer, and extended pre-training of the router. On the right side we show evaluation results using the checkpoints after 1 and 2 epochs of training.  The Noisy MOE (no training) evaluation result is the solid red line while the best result of the 4 experts alone is the dashed red line. In the 2x case, router training offers benefit on several of the tasks, not only math tasks. }
\label{fig:2xloss}
\end{figure}

 We   train the routers of the 2X MOE and the 4X MOE and examine the loss curves from the training. We also compare the results of the evaluation tests across these training paradigms. See Figures \ref{fig:4xloss} and \ref{fig:2xloss}. 
 While the loss appears to decrease more when instruct-tuning both routers and embedding layers, the results are not borne out in evaluation, shown on the right in Figures \ref{fig:4xloss} and \ref{fig:2xloss}. The Noisy MOE evaluation result is shown as a red horizontal line in each plot on the right, while the best evaluation result of the expert models in each MOE is shown as a red dashed line. 
 
 For both the 4X and the 2X MOE models, training both routers and embedding layers is significantly worse than Noisy MOE and also worse than the best expert alone. Extended pre-training is also worse than Noisy MOE. Only instruct-tuning of the routers can be seen to outperform Noisy MOE. This is notable on the math tasks GMS8K and GSM8K-COT for both the 4X and the 2X MOE, as well as on ARC-challenge in the case of the 2X MOE. We thus see that some benefit can be achieved by training the routers on a small amount of targeted data, but that such training is not needed to obtain very competitive results with the MOE.

\subsection{Ablation study on llama3-8B}
\label{sec:llama3}

We examine several variants of the methodology on a different model, llama3-8B. The experimental setup  enables a comparison with LoRA adapter-based experts as well as numerous choices for the router. The Self-MOE approach of \cite{selfmoe} is similar to but not the same as that tested here as we add a router to each FFN layer of the base model, while Self-MOE uses a single global router.  In that reference, the base models,  not the instruct-tuned models, are used for the MOE base as well as for the experts which are subsequently fine-tuned. For that reason, we do the same, but we evaluate  the instruct-tuned base model as a point of comparison. 

Specifically, we  wish to answer the following questions:
\begin{itemize}
    \item What is the impact of  the MOE base model? We thus evaluate a FLAN-trained, a medical-domain-trained, and a math-trained llama3-8B as the MOE base.
    \item How should expert models be included in the MOE? We assess using the full FFN layers, the FFN and attention layers, a fine-grained variant (called fgmlp, for 'fine-grained MLP') in which a there are three routers for each FFN layer (one each for gate, up and down), and using LoRA adapters as each expert. 
    \item Should one  train after creating the MOE? We compare the our Gate-less and Noisy MOE which do not undergo training to training the router and  training the router and  embedding layers using  llama3-8B models.
    \item How does the MOE perform compared to the most relevant baselines: the base model, the instruct-tuned base model, and the expert models alone?
\end{itemize}

The hyperparameters are as follows. The expert models were trained using num\_train\_epochs = 3, per\_device\_ train\_batch\_size = 8, gradient \_accumulation \_steps = 2, learning\_rate = 1e-06, and lr\_scheduler\_type as constant. When the routers are trained, as before, the hyperparameters used were set to num\_train\_epochs = 1, per\_device\_train\_batch\_size = 1, gradient\_accumulation \_steps = 16, learning\_rate=1e-04, and lr\_scheduler\_type as constant. 
The three experts were obtained by fine-tuning the llama3-8B base model on 50K examples each from FLAN, metamath, and  pubmedqa (pubmedqa training data only).
 When training is performed on the routers (or routers and embedding layers), in all cases the dataset used for training combines the same three datasets used metamath, FLAN, pubmedqa (pubmedqa training data only) in equal proportions of the same 50K examples each that were used to train the experts. The evaluations are performed on the same tests as above along with   medical domain tests from pubmedqa \cite{pubmedqa} and multimedqa \cite{multimedqa}.

\begin{figure}[ht]
\centering
\includegraphics[width=\linewidth]{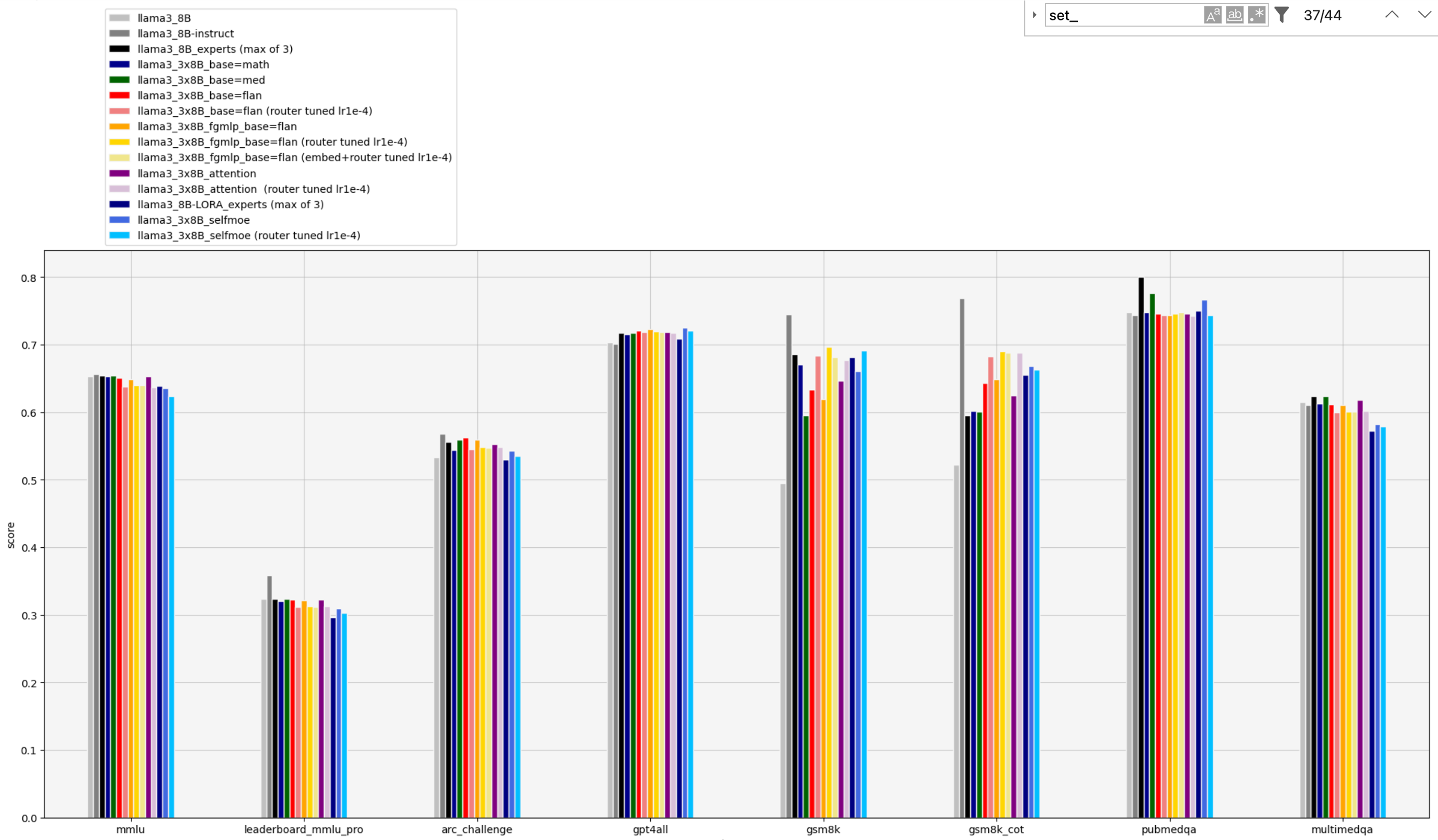}
\caption{Ablation study using several variants of our proposed methodology on llama3-8B with the relevant baselines. We evaluate the MOE with different base models, a single router per FFN layer vs. 'fgmlp' having 3 routers per FFN layer, noisy MOE, tuned router, tuned router with tuned embedding layers, and routers on FFN and attention modules. As baselines, we evaluate   llama3-8B,  instruct-tuned llama3-8B, the best fine-tuned llama3-8B for each task and the best LoRA-adapter-tuned llama3-8B for each task.  }
\label{fig:llama3}
\end{figure}

\begin{figure}[ht]
\centering
\includegraphics[width=\linewidth]{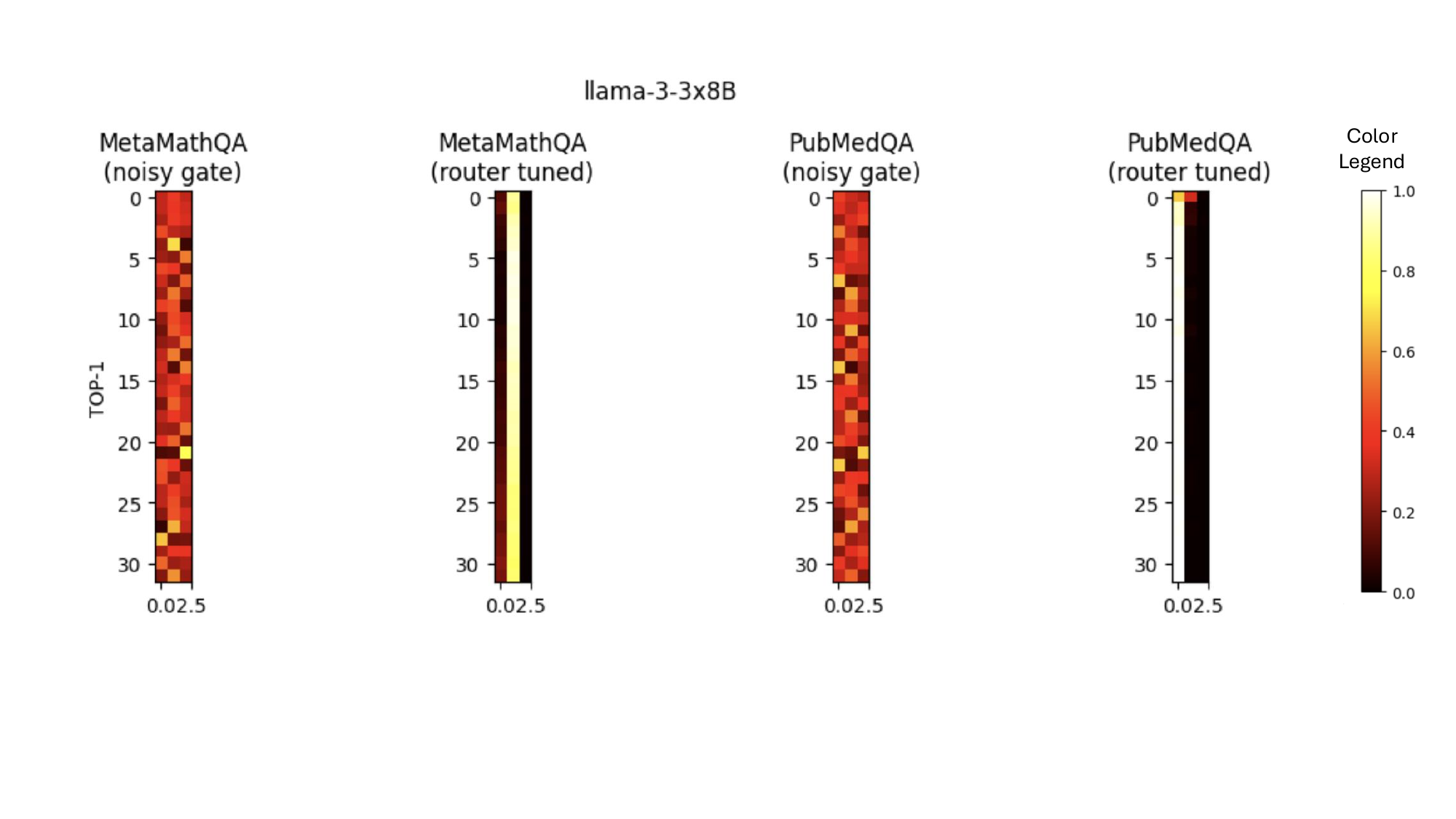}
\caption{Heat map for the llama3-8B MOE showing the percentage of time each expert is the highest-weighted expert.  In each bar there are three columns, for the FLAN, Math, and Medical experts, from left to right. The vertical axis of each column is the layer number. On the  right is the color legend used in each bar.}
\label{fig:llama3heatmap}
\end{figure}

\begin{figure}[ht]
\centering
\includegraphics[width=\linewidth]{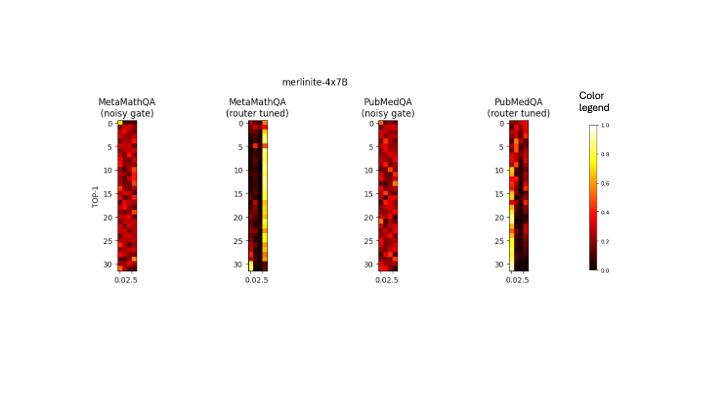}
\caption{Heat map of the Merlinite MOE showing the percentage of time each expert is the highest-weighted expert.  In each bar there are four columns, for the Merlinite, OpenChat, Hermes, and MetaMath experts, from left to right. The vertical axis of each column is the layer number. On the  right is the color legend used in each bar.}
\label{fig:merliniteheatmap}
\end{figure}

\paragraph{MOE base model has an impact on MOE performance quality} The base model used for the MOE has a noticeable impact, as can be seen from bars 4-6 (dark-blue, green and red) in Figure \ref{fig:llama3}. The MOE with a math-trained base performs the best on the GSM8K math test and the MOE with a medical-trained base performs best on the medical tests. Performance is comparable on the general knowledge and reasoning tests. For GSM8K-COT, possibly due to the importance of the question-answer format, the FLAN-instruct-trained base performs better than the MOE with the math-trained base.

\paragraph{FFN mixing  is best overall  but LoRA adapter MOE mixing is competitive}
The choice as to which modules to mix into an MOE can be seen to be application and expert-model-dependent. However, overall  FFN mixing is a better choice than mixing LoRA adapters into an MOE, as evidenced by comparing the 9 bars from bars 4-12 (dark blue to light purple) with the last two bars representing LoRA-adapter experts, (blue and cyan). 

\paragraph{Fine-grained router training can be beneficial but training is not required in general}
Comparing the pink and red bars
show that router training is not always needed though it can help performance in some cases, primarily here for the math tests, as was also the case with the Merlinite-based MOE. Comparing across the fine-grained variants (the three shades of yellow) gives the same conclusion. An interesting observation is that when the experts are LoRA adapters, contrary to the recommendation in \cite{selfmoe}, the MOE performs better when the router for the adapters is not trained. Recall  that, in these ablation tests performed on llama3-8B, the experts are fine-tuned on the same dataset used for training the routers. 
Hence, again, we see that the best recipe for creating one's own MOE will depend upon the desired use case. 

Figure \ref{fig:llama3heatmap} illustrates the  Noisy MOE and trained router weights, where the first column in each bar is the percentage of time at each layer the FLAN expert is top expert, the second column in each bar is the percentage for the math expert and the third is for the medical expert. The Noisy MOE uses a fairly even split across experts.  The router training indeed focuses attention on mainly one expert, but not always the expert one would expect to see favoured, as we see in the PubMedQA bar where it is mainly the FLAN expert that is activated as the top expert. As a point of comparison, we revisit the Merlinite MOE and show the heat map for the top expert in Figure \ref{fig:merliniteheatmap}. Note again that the router activates primarily the math expert on MetaMathQA but the medical PubMetQA favors mainly the generalist model, in this case, Merlinite.

\paragraph{Mixed MOEs can perform better than the baselines and constituent experts} The main baselines in Figure \ref{fig:llama3} are the first 3 bars: the base llama3-7B, the instruct-tuned llama3-7B, and the fine-tuned llama3-8b. We show only the best of the 3 fine-tuned expert models alone. In addition, we evaluate also the llama3-8B LoRA-tuned experts and provide the best result in the 3rd to last (very-dark-blue) bar.
Instruct-tuned llama3-7B was not used in the mixed MOEs but serves only as a point of comparison. The conclusion of this comparison is that there is often a mixed MOE that performs better than the baselines, but that a validation exercise is needed to select which is the best-performing variant.

\section{Conclusions}
\label{sec:conc}

We propose low-cost creation of an MOE from a given source model by mixing it with other expert models having the same architecture. We show that the best results are obtained when the source model is completed by well-performing models and that  a Gate-free policy, which is the cheapest option to create, is often the best or at least highly competitive with the best approach. A noisy gate can be used to reduce inference cost as compared to the Gate-free MOE, still not requiring any training, with generally  only  minor performance degradation. Both of these model mixing procedures allow for swapping in and out of expert models into an MOE at practically zero cost. We  also offer the possibility to train the routers and examine the benefits that router training provides.  As expected, results vary according to the base and expert models employed and datasets used. For that reason, the toolkit we provide the capability to use Gate-free, Noisy MOE, or router-training, and offer both FFN-based expert mixing as well as LoRA-adapter-based expert mixing. An open-source  library is provided.

\clearpage
\bibliographystyle{IEEEtran}
\bibliography{references}

\end{document}